\documentclass[11pt,a4paper]{article}
\usepackage[hyperref]{acl2018}
\usepackage{times}
\usepackage{latexsym}
\usepackage{url}

\usepackage{multirow}
\usepackage{amsmath}
\usepackage{capt-of}
\usepackage{tabularx}
\usepackage[caption=false]{subfig}
\usepackage{epsfig}
\usepackage{amssymb}
\usepackage{amsfonts}
\usepackage{booktabs}
\usepackage{scalerel}
\usepackage[inline]{enumitem}
\usepackage{listings}
\usepackage{varwidth}
\usepackage[export]{adjustbox}
\usepackage{tikz}
\usetikzlibrary{tikzmark}
\usepackage{todonotes}
\usepackage{cleveref}

\usepackage{stmaryrd}

\newcommand{\tabincell}[2]{\begin{tabular}{@{}#1@{}}#2\end{tabular}}

\DeclareMathOperator\softmax{\operatorname{softmax}}

\DeclareMathOperator\avg{\operatorname{avg}}
\DeclareMathOperator\var{\operatorname{var}}

\DeclareMathOperator\maximize{\operatorname{maximize}}

\DeclareMathOperator*{\argmax}{arg\,max}

\DeclareMathOperator*{\lstm}{\operatorname{f_{LSTM}}}

\usepackage{algorithm}
\usepackage[noend]{algpseudocode}

\definecolor{deepblue}{rgb}{0,0,0.5}
\definecolor{deepred}{rgb}{0.6,0,0}
\definecolor{deepgreen}{rgb}{0,0.5,0}
\newcommand\mybox[2][]{\tikz[overlay]\node[inner sep=1pt, anchor=text, rectangle, rounded corners=1mm,#1] {#2};\phantom{#2}}
\definecolor{fillcolor}{RGB}{216,217,252}
\newcommand\bg[1]{\mybox[fill=blue!20]{#1}}
\newcommand\rg[1]{\mybox[fill=red!20]{#1}}

\algnewcommand\algorithmicrequireb{{\hspace{0.85cm}}}
\algnewcommand\INPTDESCB{\item[\algorithmicrequireb]}

\algnewcommand\algorithmicfuncdesc{\textbf{Function:}}
\algnewcommand\FUNCDESC{\item[\algorithmicfuncdesc]}
\algnewcommand\algorithmicfuncdescb{{\hspace{1.48cm}}}
\algnewcommand\FUNCDESCB{\item[\algorithmicfuncdescb]}
\algnewcommand{\algorithmicgoto}{\textbf{goto}}
\algnewcommand{\Goto}[1]{\algorithmicgoto~\ref{#1}}
\newcommand*\Let[2]{\State {#1 $\gets$ #2}}

\newcommand*\AlgCommentInLine[1]{{\color{deepblue}{$\triangleright$ \textit{#1}}}}
\newcommand*\AlgComment[1]{\State{\AlgCommentInLine{#1}}}
\lstdefinestyle{ifttt}{
	language=Python,
	otherkeywords={Trigger,Action,-,IF,THEN},             % Add keywords here
	keywordstyle=\bfseries\color{deepblue},
	emph={MyClass,__init__},          % Custom highlighting
	emphstyle=\color{deepred},    % Custom highlighting style
	showstringspaces=false,
	breaklines=true,
	escapeinside=||,
	columns=fullflexible,
	basicstyle=\fontfamily{cmtt}\small,
	belowskip=-\baselineskip,
	aboveskip=-0.7\baselineskip
}

\lstdefinestyle{django}{
	language=Python,
	otherkeywords={self},             % Add keywords here
	keywordstyle=\bfseries\color{deepblue},
	emph={MyClass,__init__},          % Custom highlighting
	emphstyle=\color{deepred},    % Custom highlighting style
	showstringspaces=false,
	breaklines=true,
	escapeinside=||,
	columns=fullflexible,
	basicstyle=\fontfamily{cmtt}\small,
	belowskip=-\baselineskip,
	aboveskip=-0.7\baselineskip
}

\lstdefinestyle{pythoncode}{
	language=Python,
	otherkeywords={self},             % Add keywords here
	keywordstyle=\bfseries\color{deepblue},
	emph={MyClass,__init__},          % Custom highlighting
	emphstyle=\color{deepred},    % Custom highlighting style
	showstringspaces=false,
	breaklines=true,
	escapeinside=||,
	columns=fullflexible,
}

\aclfinalcopy % Uncomment this line for the final submission
\def\aclpaperid{372} %  Enter the acl Paper ID here

\title{Confidence Modeling for Neural Semantic Parsing}

\author{Li Dong$^{\dagger}$\Thanks{~Work carried out during an internship at Microsoft Research.} \and Chris Quirk$^{\ddagger}$ \and Mirella Lapata$^{\dagger}$ \\
	$^{\dagger}$ School of Informatics, University of Edinburgh \\
	$^{\ddagger}$ Microsoft Research, Redmond \\
	{\tt \href{mailto:li.dong@ed.ac.uk}{li.dong@ed.ac.uk}}~~~~{\tt \href{mailto:chrisq@microsoft.com}{chrisq@microsoft.com}}~~~~{\tt \href{mailto:mlap@inf.ed.ac.uk}{mlap@inf.ed.ac.uk}}}

\date{}

\begin{document}
\maketitle
\begin{abstract}
In this work we focus on confidence modeling for neural semantic parsers which are built upon sequence-to-sequence models. We outline three major causes of uncertainty, and design various metrics to quantify these factors. These metrics are then used to estimate confidence scores that indicate whether model predictions are likely to be correct. Beyond confidence estimation, we identify which parts of the input contribute to uncertain predictions allowing users to interpret their model, and verify or refine its input. Experimental results show that our confidence model significantly outperforms a widely used method that relies on posterior probability, and improves the quality of interpretation compared to simply relying on attention scores.
\end{abstract}

\section{Introduction}

Semantic parsing aims to map natural language text to a formal meaning
representation (e.g.,~logical forms or SQL queries). The neural
sequence-to-sequence
architecture~\cite{mt:seq2seq,mt:jointly:align:translate} has
been widely adopted in a variety of natural language processing tasks,
and semantic parsing is no exception. However, despite achieving
promising results
~\cite{lang2logic,data-recombination,latent-predictor}, neural
semantic parsers remain difficult to interpret, acting in most cases
as a black box, not providing any information about what made them
arrive at a particular decision.  In this work, we explore ways to
estimate and interpret the model's confidence in its predictions,
which we argue can provide users with immediate and meaningful
feedback regarding uncertain outputs.

An explicit framework for confidence modeling would benefit the
development cycle of neural semantic parsers which, contrary to more
traditional methods, do not make use of lexicons or templates and as a
result the sources of errors and inconsistencies are difficult to
trace.  Moreover, from the perspective of application, semantic
parsing is often used to build natural language interfaces, such as
dialogue systems. In this case it is important to know whether the
system understands the input queries with high confidence in order to
make decisions more reliably.  For example, knowing that some of the
predictions are uncertain would allow the system to generate
clarification questions, prompting users to verify the results before
triggering unwanted actions.  In addition, the training data used for
semantic parsing can be small and noisy, and as a result, models do
indeed produce uncertain outputs, which we would like our framework to
identify.

A widely-used confidence scoring method is based on posterior
probabilities~$p\left( y | x \right)$ where~$x$ is the input and
$y$~the model's prediction. For a linear model, this method makes
sense: as more positive evidence is gathered, the score becomes
larger. Neural models, in contrast, learn a complicated function that
often overfits the training data. Posterior probability is effective
when making decisions about model output, but is no longer a good
indicator of confidence due in part to the nonlinearity of neural
networks~\cite{learn-skim-read}. This observation motivates us to
develop a confidence modeling framework for sequence-to-sequence
models.  We categorize the causes of uncertainty into three types,
namely \emph{model uncertainty}, \emph{data uncertainty}, and
\emph{input uncertainty} and design different metrics to characterize
them.

We compute these confidence metrics for a given prediction and use
them as features in a regression model which is trained on held-out
data to fit prediction F1 scores. At test time, the regression model's
outputs are used as confidence scores.  Our approach does not
interfere with the training of the model, and can be thus applied to
various architectures, without sacrificing test accuracy.
Furthermore, we propose a method based on backpropagation which allows
to interpret model behavior by identifying which parts of the input
contribute to uncertain predictions. 

Experimental results on two semantic parsing datasets (\textsc{Ifttt},
\citealt{ifttt}; and \textsc{Django}, \citealt{django}) show that our
model is superior to a method based on posterior probability. We also
demonstrate that thresholding confidence scores achieves a good
trade-off between coverage and accuracy. Moreover, the proposed
uncertainty backpropagation method yields results which are
qualitatively more interpretable compared to those based on attention
scores.

\section{Related Work}
\label{sec:related-work}

\paragraph{Confidence Estimation}
Confidence estimation has been studied in the context of a few NLP
tasks, such as statistical machine
translation~\cite{mt-confidence:coling,mt-confidence:cl,mt-confidence:acl},
and question answering~\cite{watson}. To the best of our knowledge,
confidence modeling for semantic parsing remains largely unexplored.
A common scheme for modeling uncertainty in neural networks is to
place distributions over the network's
weights~\cite{nn-bayesian91,nn-bayesian92,nn-bayesian96,bbq,rnn-bayesian}. But
the resulting models often contain more parameters, and the training
process has to be accordingly changed, which makes these approaches
difficult to work with.~\newcite{dropout-bayesian} develop a
theoretical framework which shows that the use of dropout in neural
networks can be interpreted as a Bayesian approximation of Gaussian
Process. We adapt their framework so as to represent uncertainty in
the encoder-decoder architectures, and extend it by adding Gaussian
noise to weights.

\paragraph{Semantic Parsing}
Various methods have been developed to learn a semantic parser from
natural language descriptions paired with meaning
representations~\cite{tang-mooney:2000:EMNLP,zc07,lnlz08,fubl,sp:as:mt,tisp}.
More recently, a few sequence-to-sequence models have been proposed
for semantic parsing
\cite{lang2logic,data-recombination,latent-predictor} and shown to
perform competitively whilst eschewing the use of templates or
manually designed features. There have been several efforts to improve
these models including the use of a tree decoder~\cite{lang2logic},
data augmentation~\cite{data-recombination,semi-nsp}, the use of a
grammar model~\cite{grammar-nsp,asn,nl2code,table-nsp}, coarse-to-fine
decoding~\cite{coarse2fine}, network
sharing~\cite{multilingual-nsp,multi-kb-nsp}, user
feedback~\cite{user-feedback}, and transfer
learning~\cite{transfer-nsp}. Current semantic parsers will by default
generate some output for a given input even if this is just a random
guess. System results can thus be somewhat unexpected inadvertently
affecting user experience. Our goal is to mitigate these issues with a
confidence scoring model that can estimate how likely the prediction
is correct.

\section{Neural Semantic Parsing Model}
\label{sec:background}

\begin{figure}[t]
	\centering
	\includegraphics[width=0.46\textwidth]{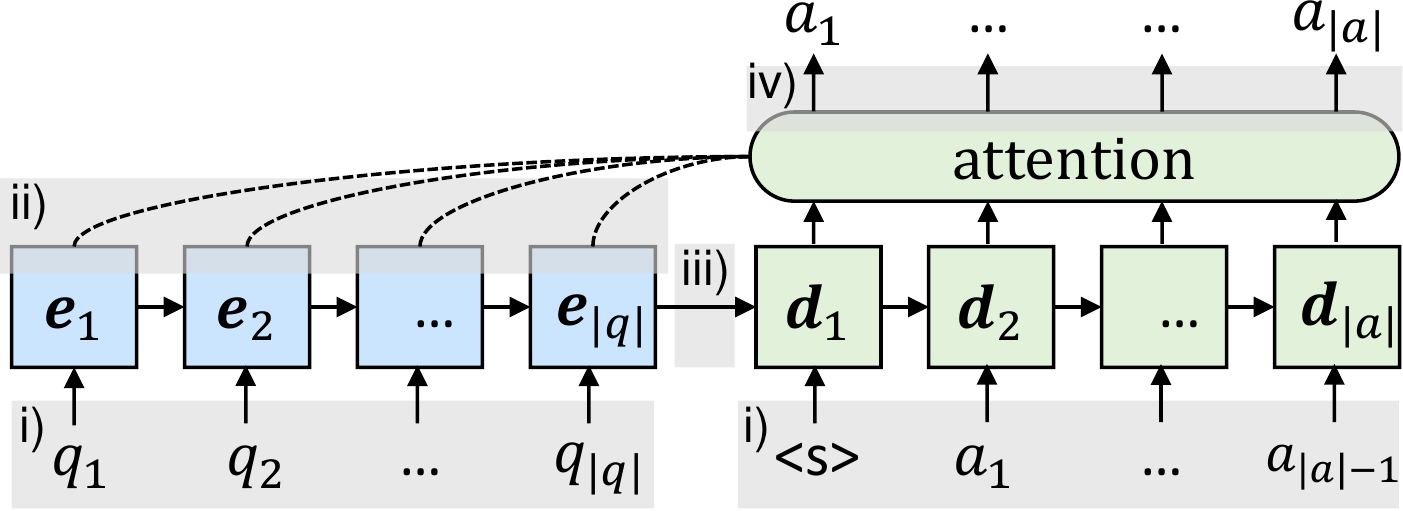}
	%\vspace{-2ex}
	\caption{We use dropout as approximate Bayesian inference to
		obtain model uncertainty. The dropout layers are applied to
		i)~token vectors; ii)~the encoder's output vectors; iii)~bridge
		vectors; and iv)~decoding vectors.}
	\label{fig:dropout_where}
\end{figure}

In the following section we describe the neural semantic parsing
model~\cite{lang2logic,data-recombination,latent-predictor} we assume
throughout this paper. The model is built upon the
sequence-to-sequence architecture and is illustrated in
Figure~\ref{fig:dropout_where}.  An \textit{encoder} is used to encode
natural language input $q = q_1 \cdots q_{|q|}$ into a vector
representation, and a \textit{decoder} learns to generate a logical
form representation of its meaning $a = a_1 \cdots a_{|a|}$
conditioned on the encoding vectors.  The encoder and decoder are two
different recurrent neural networks with long short-term memory units
(LSTMs; \citealt{lstm}) which process tokens sequentially. The
probability of generating the whole sequence $p\left( a | q \right)$
is factorized as:
\begin{equation}
\label{eq:prob:whole:seq}
p\left( a | q \right) = \prod _{ t = 1 }^{ |a| }{ p\left( a_t | a_{<t} , q \right) }
\end{equation}
where $a_{<t} = a_1 \cdots a_{t-1}$.

Let ${\mathbf{e}}_{t} \in \mathbb{R}^{n}$ denote the hidden vector of
the encoder at time step~$t$. It is computed via \mbox{${\mathbf{e}}_{t} = \lstm \left( {\mathbf{e}}_{t-1}, {\mathbf{q}}_{t} \right)$}, where
$\lstm$ refers to the LSTM unit, and
${\mathbf{q}}_{t} \in \mathbb{R}^{n}$ is the word embedding of
$q_t$. Once the tokens of the input sequence are encoded into vectors,
${\mathbf{e}}_{|q|}$ is used to initialize the hidden states of the
first time step in the decoder.

Similarly, the hidden vector of the decoder at time step~$t$ is
computed by ${\mathbf{d}}_{t} = \lstm \left( {\mathbf{d}}_{t-1},
  {\mathbf{a}}_{t-1} \right)$, where ${\mathbf{a}}_{t-1} \in
\mathbb{R}^{n}$ is the word vector of the previously predicted
token. Additionally, we use an attention
mechanism~\cite{luong-attention} to utilize relevant encoder-side
context. For the current time step $t$ of the decoder, we compute its
attention score with the \mbox{$k$-th}~hidden state in the encoder as:
\begin{equation}
\label{eq:attention:score}
{ r }_{ t,k } \propto { \exp \{ {\mathbf{d}}_{ t } \cdot {\mathbf{e}}_{ k } \} }
\end{equation}
where $\sum _{ j=1 }^{ |q| }{ { r }_{ t,j } } =1$.
The probability of generating $a_t$ is computed via:
\begin{align}
\mathbf{c}_{t} &= \sum_{ k=1 }^{ |q| }{ { r }_{ t,k } {\mathbf{e}}_{ k } } \\
{\mathbf{d} }_{ t }^{ att } &= \tanh \left( \mathbf{W}_1 {\mathbf{d} }_{ t } + \mathbf{W}_2 \mathbf{c}_{t} \right) \label{eq:attention:new:hidden} \\
p\left( a_t | a_{<t} , q \right) &= \softmax_{a_t} \left( \mathbf{W}_o {\mathbf{d} }_{ t }^{ att } \right) \label{eq:attention:decoder:predict}
\end{align}
where $\mathbf{W}_1 , \mathbf{W}_2 \in \mathbb{R}^{n \times n}$ and $\mathbf{W}_o \in \mathbb{R}^{|V_a| \times n}$ are three parameter matrices.

The training objective is to maximize the likelihood of the generated
meaning representation~$a$ given input~$q$, i.e., $\maximize \sum_{(q
  , a) \in \mathcal{D} }{ \log{p \left( a | q \right)}}$, where
$\mathcal{D}$ represents training pairs. At test time, the model's
prediction for input~$q$ is obtained via $\hat{a} = \argmax_{a'}{ p
  \left( a' | q \right) }$, where $a'$ represents candidate
outputs. Because $p\left( a | q \right)$ is factorized as shown in
Equation~\eqref{eq:prob:whole:seq}, we can use beam search to generate
tokens one by one rather than iterating over all possible results.

\section{Confidence Estimation}
\label{sec:method:estimation}

Given input~$q$ and its predicted meaning representation~$a$, the
confidence model estimates score~$s\left(q,a\right) \in (0,1)$. A
large score indicates the model is confident that its prediction is
correct. In order to gauge confidence, we need to estimate ``what we
do not know''. To this end, we identify three causes of uncertainty,
and design various metrics characterizing each one of them. We then
feed these metrics into a regression model in order to predict
$s\left(q,a\right)$.

\subsection{Model Uncertainty}
\label{sec:method:metric-model}

The model's parameters or structures contain uncertainty, which makes
the model less confident about the values of $p\left( a | q
\right)$. For example, noise in the training data and the stochastic
learning algorithm itself can result in model uncertainty.  We
describe metrics for capturing uncertainty below:

\begin{algorithm}[t]
\caption{Dropout Perturbation
\label{alg:dropout}}
\begin{algorithmic}[1]
\small
\Require $q, a$: Input and its prediction
\INPTDESCB $\mathcal{M}$: Model parameters

\For{$i \gets 1,\cdots,F$}
\Let{$\hat{\mathcal{M}}^{i}$}{Apply dropout layers to $\mathcal{M}$}\hfill\AlgCommentInLine{Figure~\ref{fig:dropout_where}}
\State{Run forward pass and compute $\hat{p}( a | q ; \hat{\mathcal{M}}^{i} )$}
\EndFor
\State{Compute variance of $\{ \hat{p}( a | q ; \hat{\mathcal{M}}^{i} ) \}_{i=1}^{F}$}\hfill\AlgCommentInLine{Equation~\eqref{eq:dropout-uncertainty-seq}}
\normalsize
\end{algorithmic}
\end{algorithm}

\paragraph{Dropout Perturbation}
Our first metric uses dropout~\cite{dropout} as approximate Bayesian
inference to estimate model
uncertainty~\cite{dropout-bayesian}. Dropout is a widely used
regularization technique during training, which relieves overfitting
by randomly masking some input neurons to zero according to a
Bernoulli distribution. In our work, we use dropout at \emph{test
  time}, instead. As shown in Algorithm~\ref{alg:dropout},
we perform $F$~forward passes through the network,
and collect the results $\{ \hat{p}( a | q ; \hat{\mathcal{M}}^{i} )
\}_{i=1}^{F}$ where $\hat{\mathcal{M}}^{i}$ represents the perturbed
parameters. Then, the uncertainty metric is computed by the variance
of results. We define the metric on the sequence level as:
\begin{equation}
\label{eq:dropout-uncertainty-seq}
\var \{ \hat{p}( a | q ; \hat{\mathcal{M}}^{i} ) \}_{i=1}^{F} \text{.}
\end{equation}
In addition, we compute uncertainty~$u_{a_t}$ at the token-level~$a_t$
via:
\begin{equation}
\label{eq:token-uncertainty}
u_{a_t} = \var \{ \hat{p}( a_t | a_{<t} , q ; \hat{\mathcal{M}}^{i} ) \}_{i=1}^{F}
\end{equation}
where $\hat{p}( a_t | a_{<t} , q ; \hat{\mathcal{M}}^{i} )$ is the
probability of generating token~$a_t$
(Equation~\eqref{eq:attention:decoder:predict}) using perturbed
model~$\hat{\mathcal{M}}^{i}$.  We operationalize token-level
uncertainty in two ways, as the average score $\avg \{ u_{a_t}
\}_{t=1}^{|a|}$ and the maximum score $\max \{ u_{a_t} \}_{t=1}^{|a|}$
(since the uncertainty of a sequence is often determined by the most
uncertain token).
As shown in Figure~\ref{fig:dropout_where}, we add dropout layers in
i)~the word vectors of the encoder and decoder $\mathbf{q}_{t} ,
\mathbf{a}_{t}$; ii)~the output vectors of the encoder
$\mathbf{e}_{t}$; iii)~bridge vectors~${\mathbf{e}}_{|q|}$ used to
initialize the hidden states of the first time step in the decoder;
and iv)~decoding vectors ${\mathbf{d} }_{ t }^{ att }$ (Equation~\eqref{eq:attention:new:hidden}).  

\paragraph{Gaussian Noise}
Standard dropout can be viewed as applying noise sampled from a
Bernoulli distribution to the network parameters. We instead use
Gaussian noise, and apply the metrics in the same way discussed
above. Let~$\mathbf{v}$ denote a vector perturbed by noise, and
$\mathbf{g}$~a vector sampled from the Gaussian
distribution~$\mathcal{N}(0,\sigma^2)$. We use $\hat{\mathbf{v}} =
\mathbf{v}+\mathbf{g}$ and $\hat{\mathbf{v}} = \mathbf{v}+\mathbf{v}
\odot \mathbf{g}$ as two noise injection methods. Intuitively, if the
model is more confident in an example, it should be more robust to
perturbations.

\paragraph{Posterior Probability}
Our last class of metrics is based on posterior probability.  We use
the log probability $\log p(a|q)$ as a sequence-level metric.  The
token-level metric $\min \{ p( a_t | a_{<t} , q) \}_{t=1}^{|a|}$ can
identify the most uncertain predicted token.  The perplexity per
token~$-\frac{1}{|a|}\sum_{ t = 1 }^{ |a| }{ \log{ p\left( a_t |
      a_{<t} , q \right) } }$ is also employed.

\subsection{Data Uncertainty}
\label{sec:method:metric-data}

The coverage of training data also affects the uncertainty of
predictions. If the input $q$ does not match the training distribution
or contains unknown words, it is difficult to predict~$p\left( a | q
\right)$ reliably. We define two metrics:

\paragraph{Probability of Input}
We train a language model on the training data, and use it to estimate
the probability of input $p (q|\mathcal{D})$ where $\mathcal{D}$
represents the training data.

\paragraph{Number of Unknown Tokens}
Tokens that do not appear in the training data harm robustness, and
lead to uncertainty. So, we use the number of unknown tokens in the
input~$q$ as a metric.

\subsection{Input Uncertainty}
\label{sec:method:metric-input}

Even if the model can estimate $p\left( a | q \right)$ reliably, the
input itself may be ambiguous. For instance, the input \textsl{the
  flight is at 9 o'clock} can be interpreted as either
\lstinline[style=ifttt]|flight_time(9am)| or
\lstinline[style=ifttt]|flight_time(9pm)|. Selecting between these
predictions is difficult, especially if they are both highly
likely. We use the following metrics to measure uncertainty caused by
ambiguous inputs.

\paragraph{Variance of Top Candidates}
We use the variance of the probability of the top candidates to
indicate whether these are similar. The sequence-level metric is
computed by:
\begin{equation}
\var \{ p( a^{i} | q ) \}_{i=1}^{K} \nonumber
\end{equation}
where $a^{1} \dots a^{K}$
are the $K$-best predictions obtained by the beam search during
inference (Section~\ref{sec:background}).

\paragraph{Entropy of Decoding}
The sequence-level entropy of the decoding process is computed via:
\begin{equation}
H[a|q] = -\sum_{a'} {p(a'|q) \log p(a'|q)} \nonumber
\end{equation}
which we approximate by
Monte Carlo sampling rather than iterating over all candidate predictions.
The token-level metrics of decoding entropy are
computed by $\avg \{ H[ a_t | a_{<t} , q] \}_{t=1}^{|a|}$ and $\max \{
H[ a_t | a_{<t} , q] \}_{t=1}^{|a|}$.

\subsection{Confidence Scoring}
\label{sec:method:scoring}

The sentence- and token-level confidence metrics defined in
Section~\ref{sec:method:estimation} are fed into a
gradient tree boosting model~\cite{xgboost} in order to predict the
overall confidence score~$s\left(q,a\right)$. The model is wrapped
with a logistic function so that confidence scores are in the range
of~$(0,1)$.

Because the confidence score indicates whether the prediction is
likely to be correct, we can use the prediction's F1 (see
Section~\ref{sec:setting}) as target value. The training loss is
defined as:
\begin{equation}
\sum_{(q,a) \in
\mathcal{D}}{\hspace*{-1.2ex}\ln(1\hspace*{-.5ex}+\hspace*{-.5ex} e^{-\hat{s}(q,a)})^{y_{q,a}} \hspace*{-.5ex}+ \ln(1 \hspace*{-.5ex}+\hspace*{-.5ex}e^{\hat{s}(q,a)})^{(1 - y_{q,a})} } \nonumber
\end{equation}
where $\mathcal{D}$ represents the data, $y_{q,a}$ is the
target F1 score, and $\hat{s}(q,a)$ the predicted confidence score.
We refer readers to~\citet{xgboost} for mathematical details of how
the gradient tree boosting model is trained.
Notice that we learn the confidence scoring model on the
held-out set (rather than on the training data of the semantic parser)
to avoid overfitting.

\section{Uncertainty Interpretation}
\label{sec:method:interpret}

Confidence scores are useful in so far they can be traced back to the
inputs causing the uncertainty in the first place. For semantic
parsing, identifying which input words contribute to uncertainty would
be of value, e.g.,~these could be treated explicitly as special
cases or refined if they represent noise.

\begin{figure}[t]
	\centering
	\includegraphics[width=0.48\textwidth]{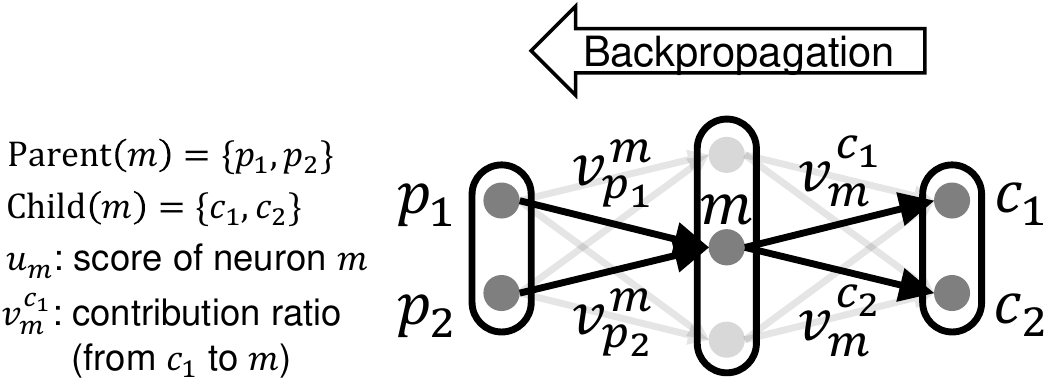}
%	\vspace{-3ex}
	\caption{Uncertainty backpropagation at the neuron
		level. Neuron $m$'s score $u_m$ is collected from  child
		neurons $c_1$ and $c_2$ by $u_m = v_{m}^{c_1} u_{c_1} +
		v_{m}^{c_2} u_{c_2}$. The score $u_m$ is then redistributed
		to its parent neurons $p_1$ and $p_2$, which satisfies
		$v_{p_1}^{m} + v_{p_2}^{m} = 1$.}
	\label{fig:uncertainty_bp}
\end{figure}

In this section, we introduce an algorithm that backpropagates
token-level uncertainty scores (see
Equation~\eqref{eq:token-uncertainty}) from predictions to input
tokens, following the ideas of~\newcite{lrp} and \newcite{excitation}.
Let $u_{m}$ denote neuron~$m$'s uncertainty score, which indicates the
degree to which it contributes to uncertainty.  As shown in
Figure~\ref{fig:uncertainty_bp}, $u_{m}$ is computed by the summation
of the scores backpropagated from its child neurons:
\begin{equation}
u_{m} = \sum_{c \in \textrm{Child}(m)} {v_{m}^{c} u_{c}} \nonumber
\end{equation}
where $\textrm{Child}(m)$ is the set of $m$'s child neurons, and the
non-negative contribution ratio $v_{m}^{c}$ indicates how much we
backpropagate $u_{c}$ to neuron~$m$.  Intuitively, if neuron~$m$
contributes more to $c$'s value, ratio~$v_{m}^{c}$ should be
larger.

After obtaining score~$u_{m}$, we redistribute it to its parent
neurons in the same way.  Contribution ratios from~$m$ to its parent
neurons are normalized to~$1$:
\begin{equation}
\sum_{p \in \textrm{Parent}(m)} {v_{p}^{m}} = 1 \nonumber
\end{equation}
where $\textrm{Parent}(m)$ is the set of $m$'s parent neurons.

Given the above constraints, we now define different backpropagation
rules for the operators used in neural networks.  We first describe
the rules used for fully-connected layers. Let $\mathbf{x}$ denote the
input. The output is computed by $\mathbf{z} = \sigma ( \mathbf{W}
\mathbf{x} + \mathbf{b} )$, where $\sigma$ is a nonlinear function,
$\mathbf{W} \in \mathbb{R}^{|\mathbf{z}|*|\mathbf{x}|}$ is the weight
matrix, $\mathbf{b} \in \mathbb{R}^{|\mathbf{z}|}$ is the bias, and
neuron~$\mathbf{z}_{i}$ is computed via $\mathbf{z}_{i} = \sigma (
\sum_{j=1}^{|\mathbf{x}|}{\mathbf{W}_{i,j} \mathbf{x}_{j}} +
\mathbf{b}_i )$. Neuron~$\mathbf{x}_k$'s uncertainty score $u_{x_k}$
is gathered from the next layer:
\begin{equation}
u_{x_k} = \sum_{i=1}^{|\mathbf{z}|}{v_{x_k}^{z_i} u_{z_i}} = \sum_{i=1}^{|\mathbf{z}|}{ \frac{|\mathbf{W}_{i,k} \mathbf{x}_{k}|}{\sum_{j=1}^{|\mathbf{x}|}{|\mathbf{W}_{i,j} \mathbf{x}_{j}|}} u_{z_i}} \nonumber
\end{equation}
ignoring the nonlinear function~$\sigma$ and the bias~$\mathbf{b}$.
The ratio $v_{x_k}^{z_i}$ is proportional to the contribution of $\mathbf{x}_k$ to the value of $\mathbf{z}_i$.

We define backpropagation rules for element-wise vector operators. For
$\mathbf{z} = \mathbf{x} \pm \mathbf{y}$, these are:

\begin{equation}
\begin{array}{ll}
u_{x_k} = \frac{|\mathbf{x}_k|}{|\mathbf{x}_k| + |\mathbf{y}_k|} u_{z_k} & 
u_{y_k} = \frac{|\mathbf{y}_k|}{|\mathbf{x}_k| + |\mathbf{y}_k|} u_{z_k}
\end{array} \nonumber
\end{equation}
where the contribution ratios $v_{x_k}^{z_k}$ and $v_{y_k}^{z_k}$ are
determined by $|\mathbf{x}_k|$ and $|\mathbf{y}_k|$.  For
multiplication, the contribution of two elements in $\frac{1}{3} * 3$
should be the same. So, the propagation rules for $\mathbf{z} =
\mathbf{x} \odot \mathbf{y}$ are:
\begin{equation}
\hspace*{-.73cm}\begin{array}{@{~}l@{~}l@{~}}
u_{x_k} \hspace*{-.6ex}=\hspace*{-.6ex} \frac{|\log {|\mathbf{x}_k|} |}{|\log {|\mathbf{x}_k|} | + |\log
	{|\mathbf{y}_k|}| } u_{z_k} & 
u_{y_k} \hspace*{-.6ex}=\hspace*{-.6ex} \frac{|\log {|\mathbf{y}_k|} |}{|\log {|\mathbf{x}_k|} | + |\log {|\mathbf{y}_k|} |} u_{z_k}\\
\end{array}\hspace*{-.6cm} \nonumber
\end{equation}
where the contribution ratios are determined by $|\log {|\mathbf{x}_k|} |$ and $|\log {|\mathbf{y}_k|} |$.

For scalar multiplication, $\mathbf{z} = \lambda \mathbf{x}$ where
$\lambda$ denotes a constant. We directly assign~$\mathbf{z}$'s
uncertainty scores to~$\mathbf{x}$ and the backpropagation rule is
$u_{x_k} = u_{z_k}$.

%%%%%%%%%%%%%%%%%%%%%%%%%%%%%%%%%%%%%%%%%%%%%%

\begin{algorithm}[t]
\caption{Uncertainty Interpretation
\label{alg:bp}}
\begin{algorithmic}[1]
\small
\Require $q, a$: Input and its prediction
\Ensure $\{\hat{u}_{q_t}\}_{t=1}^{|q|}$: Interpretation scores for input tokens
\FUNCDESC $\mathsf{TokenUnc}$: Get token-level uncertainty

\AlgComment{Get token-level uncertainty for predicted tokens} \label{alg:line:init}
\Let{$\{u_{a_t}\}_{t=1}^{|a|}$}{$\mathsf{TokenUnc}(q,a)$}
\AlgComment{Initialize uncertainty scores for backpropagation}
\For{$t \gets 1,\cdots,|a|$}
\Let{Decoder classifier's output neuron}{$u_{a_t}$}
\EndFor \label{alg:line:init:end}
\AlgComment{Run backpropagation} \label{alg:line:bp}
\For{$m \gets$ neuron in backward topological order}
\AlgComment{Gather scores from child neurons}
\Let{$u_{m}$}{$\sum_{c \in \textrm{Child}(m)} {v_{m}^{c} u_{c}}$}
\EndFor \label{alg:line:bp:end}
\AlgComment{Summarize scores for input words} \label{alg:line:summ}
\For{$t \gets 1,\cdots,|q|$}
\Let{$u_{q_t}$}{$\sum_{c \in \mathbf{q}_t} {u_{c}}$}
\EndFor
\Let{$\{\hat{u}_{q_t}\}_{t=1}^{|q|}$}{normalize $\{u_{q_t}\}_{t=1}^{|q|}$} \label{alg:line:summ:end}
\normalsize
\end{algorithmic}
\end{algorithm}

\begin{table*}[t]
\centering
\small
\begin{tabular}{p{1.2cm} p{13.4cm}}
\toprule
\textbf{Dataset} & \textbf{Example} \\ \midrule
\multirow{2}{*}{\textsc{Ifttt}} & \textit{turn android phone to full volume at 7am monday to friday} \\
&
\begin{lstlisting}[style=ifttt,basicstyle=\fontfamily{cmtt}\footnotesize]
date_time-every_day_of_the_week_at-((time_of_day (07)(:)(00)) (days_of_the_week (1)(2)(3)(4)(5))) THEN android_device-set_ringtone_volume-(volume ({'volume_level':1.0,'name':'100%'}))
\end{lstlisting}
\\ \midrule
\multirow{2}{*}{\textsc{Django}} & \textit{for every key in sorted list of user\_settings} \\
&
\begin{lstlisting}[style=django]
for key in sorted(user_settings):
\end{lstlisting}
\\ \bottomrule
\end{tabular}
\normalsize
%\vspace{-1ex}
\caption{Natural language descriptions and their meaning representations from \textsc{Ifttt} and \textsc{Django}.}
\label{table:dataset}
\end{table*}

As shown in Algorithm~\ref{alg:bp}, we first initialize uncertainty
backpropagation in the decoder
(lines~\mbox{\ref{alg:line:init}--\ref{alg:line:init:end}}).  For each
predicted token $a_t$, we compute its uncertainty score $u_{a_t}$ as
in Equation~\eqref{eq:token-uncertainty}. Next, we find the dimension
of~$a_t$ in the decoder's softmax classifier
(Equation~\eqref{eq:attention:decoder:predict}), and initialize the
neuron with the uncertainty score $u_{a_t}$.  We then backpropagate
these uncertainty scores through the network
(lines~\mbox{\ref{alg:line:bp}--\ref{alg:line:bp:end}}),
%\subsection{Score Summarization}
and finally into the neurons of the input words. We summarize them and
compute the token-level scores for interpreting the results
(line~\ref{alg:line:summ}--\ref{alg:line:summ:end}).  For input word
vector $\mathbf{q}_{t}$, we use the summation of its neuron-level
scores as the token-level score:
\begin{equation}
\hat{u}_{q_t} \propto \sum_{c \in \mathbf{q}_t} {u_{c}} \nonumber
\end{equation}
where $c \in \mathbf{q}_t$ represents the neurons of word vector $\mathbf{q}_t$, and $\sum _{ t=1 }^{ |q| }{ \hat{u}_{q_t} } = 1$.
We use the normalized score~$\hat{u}_{q_t}$ to indicate token
$q_t$'s contribution to prediction uncertainty.

\section{Experiments}
\label{sec:exp}

In this section we describe the datasets used in our experiments and
various details concerning our models. We present our experimental
results and analysis of model behavior. Our code is publicly available
at \url{https://github.com/donglixp/confidence}.

\subsection{Datasets}
\label{sec:exp:dataset}

We trained the neural semantic parser introduced in
Section~\ref{sec:background} on two datasets covering different
domains and meaning representations.  Examples are shown in
Table~\ref{table:dataset}.

\paragraph{\textsc{Ifttt}}
This dataset~\cite{ifttt} contains a large number of if-this-then-that
programs crawled from the \textsc{Ifttt}
website.
%\footnote{\url{http://ifttt.com}}
The programs
are written for various
applications, such as home security (e.g.,~``\textsl{email me if the
	window opens}''), and task automation (e.g.,~``\textsl{save
	instagram photos to dropbox}''). Whenever a program's trigger is
satisfied, an action is performed. Triggers and actions represent
functions with arguments; they are selected from different channels
($160$ in total)
representing various services (e.g., Android).
There are~$552$~trigger functions and~$229$~action functions.
The original split contains~$77,495$ training,
$5,171$~development, and $4,294$~test instances.  The subset that
removes non-English descriptions was used in our experiments.

\paragraph{\textsc{Django}}
This dataset~\cite{django} is built upon the code of the Django web
framework. Each line of Python code has a manually annotated natural
language description. Our goal is to map the English pseudo-code to
Python statements. This dataset contains diverse use cases, such as
iteration, exception handling, and string manipulation. The original
split has $16,000$ training, $1,000$ development, and $1,805$ test
examples.

\subsection{Settings}
\label{sec:setting}

We followed the data preprocessing used in previous work~\cite{lang2logic,nl2code}.
Input sentences were tokenized using NLTK~\cite{nltk} and lowercased.
We filtered words that appeared less than four times in the training set.
Numbers and URLs in
\textsc{Ifttt} and quoted strings in \textsc{Django} were replaced
with place holders.  Hyperparameters of the semantic
parsers were validated on the
development set.
The learning rate and the smoothing constant of RMSProp~\cite{rmsprop} were~$0.002$ and~$0.95$, respectively.
The dropout rate was $0.25$. A two-layer LSTM was used for \textsc{Ifttt}, while a
one-layer LSTM was employed for \textsc{Django}. Dimensions for
the word embedding and hidden vector were selected from
$\{150,250\}$. The beam size during decoding was~$5$.

For \textsc{Ifttt}, we view the predicted trees as a set of
productions, and use balanced F1 as evaluation metric~\cite{ifttt}.
We do not measure accuracy because the dataset is very noisy and there
rarely is an exact match between the predicted output and the gold
standard.  The F1 score of our neural semantic parser is $50.1\%$,
which is comparable to~\newcite{lang2logic}.  For \textsc{Django}, we
measure the fraction of exact matches, where F1 score is equal to
accuracy. Because there are unseen variable names at test time, we use
attention scores as alignments to replace unknown tokens in the
prediction with the input words they align
to~\cite{mt:rare:word:google}. The accuracy of our parser is $53.7\%$,
which is better than the result ($45.1\%$) of the sequence-to-sequence
model reported in~\newcite{nl2code}.

To estimate model uncertainty, we set dropout rate to $0.1$, and
performed $30$ inference passes. The standard deviation of Gaussian
noise was $0.05$.  The language model was estimated using
KenLM~\cite{kenlm}.  For input uncertainty, we computed variance for
the $10$-best candidates.
The confidence metrics were implemented in batch mode, to take full advantage of GPUs.
Hyperparameters of the confidence scoring model were cross-validated.
The number of boosted trees was selected from $\{20,50\}$.
The maximum tree depth was selected from
$\{3,4,5\}$. We set the subsample ratio to~$0.8$. All other
hyperparameters in XGBoost~\cite{xgboost} were left with their
default values.

\subsection{Results}

\begin{table}[t]
\centering
\small
\begin{tabular}{l c c}
\toprule
\textbf{Method}      & \textbf{\textsc{Ifttt}} & \textbf{\textsc{Django}} \\ \midrule
\textsc{Posterior}   & 0.477                   & 0.694                    \\ \midrule
\textsc{Conf}        & \textbf{0.625}          & \textbf{0.793}           \\
~~$-$ \textsc{Model} & 0.595                   & 0.759                    \\
~~$-$ \textsc{Data}  & 0.610                   & 0.787                    \\
~~$-$ \textsc{Input} & 0.608                   & 0.785                    \\ \bottomrule
\end{tabular}
\normalsize
\caption{Spearman~$\rho$ correlation between confidence
scores and F1. Best results are shown in \textbf{bold}. All
correlations are significant at \mbox{$p<0.01$}.} 
	\label{table:results:estimation}
\end{table}

\paragraph{Confidence Estimation}
\label{sec:exp:estimation}
We compare our approach (\textsc{Conf}) against confidence scores
based on posterior probability~$p(a|q)$ (\textsc{Posterior}). We also
report the results of three ablation variants ($-$\textsc{Model},
$-$\textsc{Data}, $-$\textsc{Input}) by removing each group of
confidence metrics described in
Section~\ref{sec:method:estimation}. We measure the relationship
between confidence scores and F1 using Spearman's~$\rho$ correlation
coefficient which varies between $-1$ and $1$ ($0$~implies there is no
correlation). High~$\rho$ indicates that the confidence scores are
high for correct predictions and low otherwise.

As shown in Table~\ref{table:results:estimation}, our method
\textsc{Conf} outperforms \textsc{Posterior} by a large margin.  The
ablation results indicate that model uncertainty plays the most
important role among the confidence metrics. In contrast, removing the
metrics of data uncertainty affects performance less, because most
examples in the datasets are in-domain. Improvements for each group of
metrics are significant with $p<0.05$ according to bootstrap
hypothesis testing~\cite{bootstrap}.

\begin{table}[t]
	\small
	\begin{center}
		\begin{tabular}{@{}l@{~~}c@{~~}c@{~~}c@{~~}c@{~~}c@{~~}c@{~~}c@{~~}c@{~}}
			\toprule
			& F1   & Dout & Noise & PR   & PPL & LM   & \#UNK & Var  \\ \midrule
			Dout  & \textbf{0.59} &      &       &      & &      &       &  \\
			Noise & \textbf{0.59} & 0.90 &       &      & &      &       &  \\
			PR    & \textbf{0.52} & 0.84 & 0.82  &      & &      &       &  \\
			PPL   & 0.48 & 0.78 & 0.78  & 0.89 & &      &       &  \\
			LM    & 0.30 & 0.26 & 0.32  & 0.27 & 0.25 &      &       &  \\
			\#UNK & 0.27 & 0.31 & 0.33  & 0.29 & 0.25 & 0.32 &       &  \\
			Var   & 0.49 & 0.83 & 0.78  & 0.88 & 0.79 & 0.25 & 0.27  &  \\
			Ent   & 0.53 & 0.78 & 0.78  & 0.80 & 0.75 & 0.27 & 0.30  & 0.76 \\ \bottomrule
		\end{tabular}
	\end{center}
	\normalsize
	%\vspace{-1ex}
	\caption{Correlation matrix for F1 and individual confidence metrics
		on the \textsc{Ifttt} dataset. All correlations are significant at
		\mbox{$p<0.01$}. Best predictors are shown in \textbf{bold}. Dout is short for dropout, PR for posterior
		probability, PPL for perplexity, LM for probability based on a
		language model, \#UNK for number of unknown tokens, Var for
		variance of top candidates, and Ent for Entropy. \label{tab:corr1}}
\end{table}

\begin{table}[t]
	\small
	\begin{center}
		\begin{tabular}{@{}l@{~~}c@{~~}c@{~~}c@{~~}c@{~~}c@{~~}c@{~~}c@{~~}c@{~}}
			\toprule
			&  F1  & Dout & Noise &  PR  & PPL &  LM  & \#UNK & Var  \\ \midrule
			Dout  & \textbf{0.76} &      &       &      & &      &       &  \\
			Noise & \textbf{0.78} & 0.94 &       &      & &      &       &  \\
			PR    & \textbf{0.73} & 0.89 & 0.90  &      & &      &       &  \\
			PPL   & 0.64 & 0.80 & 0.81  & 0.84 & &      &       &  \\
			LM    & 0.32 & 0.41 & 0.40  & 0.38 & 0.30 &      &       &  \\
			\#UNK & 0.27 & 0.28 & 0.28  & 0.26 & 0.19 & 0.35 &       &  \\
			Var   & 0.70 & 0.87 & 0.87  & 0.89 & 0.87 & 0.37 & 0.23  &  \\
			Ent   & 0.72 & 0.89 & 0.90  & 0.92 & 0.86 & 0.38 & 0.26  & 0.90 \\ \bottomrule
		\end{tabular}
	\end{center}
	\normalsize
	%\vspace{-1ex}
	\caption{Correlation matrix for F1 and individual confidence metrics
		on the  \textsc{Django} dataset. All correlations are significant at
		\mbox{$p<0.01$}. Best predictors are shown in \textbf{bold}. Same shorthands apply as in  Table~\ref{tab:corr1}.\label{tab:corr2}} 
\end{table}

\begin{table}[t]
	\small
	\begin{center}
		\begin{tabular}{@{}l@{~~}c@{~~}c@{~~}c@{~~}c@{~~}c@{~~}c@{~~}c@{~~}c@{~}}
			\toprule
			\textbf{Metric} & Dout & Noise & PR   & PPL & LM   & \#UNK & Var & Ent \\ \midrule
			\textbf{\textsc{Ifttt}}  & 0.39 & \textbf{1.00}  & \textbf{0.89} & 0.27 & 0.26 & 0.46 & 0.43 & 0.34 \\
			\textbf{\textsc{Django}} & \textbf{1.00} & \textbf{0.59}  & 0.22 & \textbf{0.58} & 0.49 & 0.14 & 0.24 & 0.25 \\ \bottomrule
		\end{tabular}
	\end{center}
	\normalsize
	%\vspace{-1ex}
	\caption{Importance scores of confidence metrics (normalized by
		maximum value on each dataset). Best results are shown in \textbf{bold}. Same shorthands apply as in Table~\ref{tab:corr1}.\label{tab:metric-importance}}
\end{table}

Tables~\ref{tab:corr1} and~\ref{tab:corr2} show the correlation matrix
for F1 and individual confidence metrics on the \textsc{Ifttt} and
\textsc{Django} datasets, respectively. As can
be seen, metrics representing model uncertainty and input uncertainty
are more correlated to each other compared with metrics capturing data
uncertainty. Perhaps unsurprisingly metrics of the same group are
highly inter-correlated since they model the same type of uncertainty.
Table~\ref{tab:metric-importance} shows the relative importance of
individual metrics in the regression model. As importance score we use
the average gain (i.e., loss reduction) brought by the confidence
metric once added as feature to the branch of the decision tree
\cite{xgboost}. The results indicate that model uncertainty
(Noise/Dropout/Posterior/Perplexity) plays the most important role. On
\textsc{Ifttt}, the number of unknown tokens (\#UNK) and the variance
of top candidates (var(K-best)) are also very helpful because this
dataset is relatively noisy and contains many ambiguous inputs.

\begin{figure}[t]
	\centering
	\subfloat[\textsc{Ifttt} \label{fig:threshold:ifttt}]{%
		\includegraphics[width=0.38\textwidth]{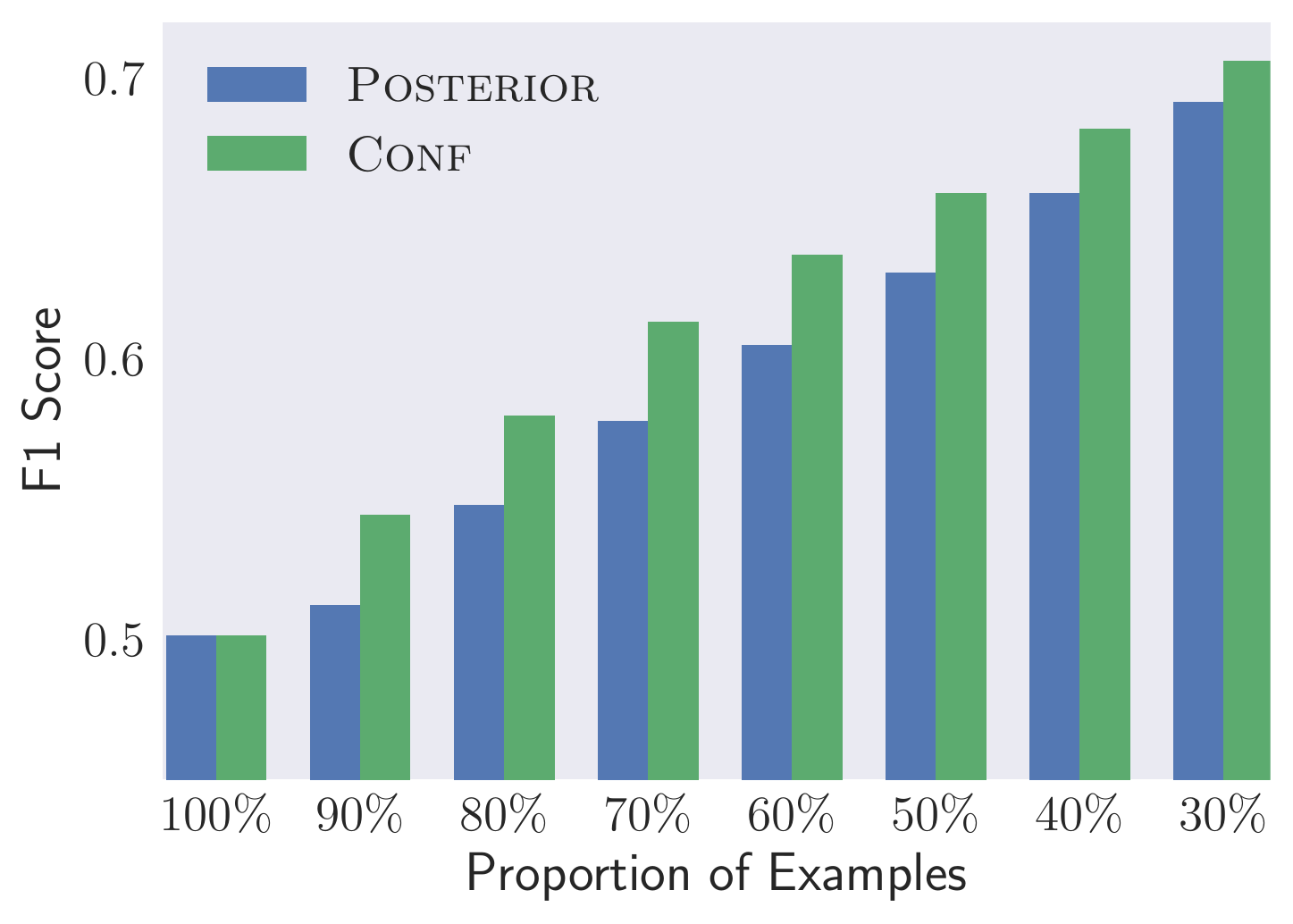}
	} \\
	\subfloat[\textsc{Django} \label{fig:threshold:django}]{%
		\includegraphics[width=0.38\textwidth]{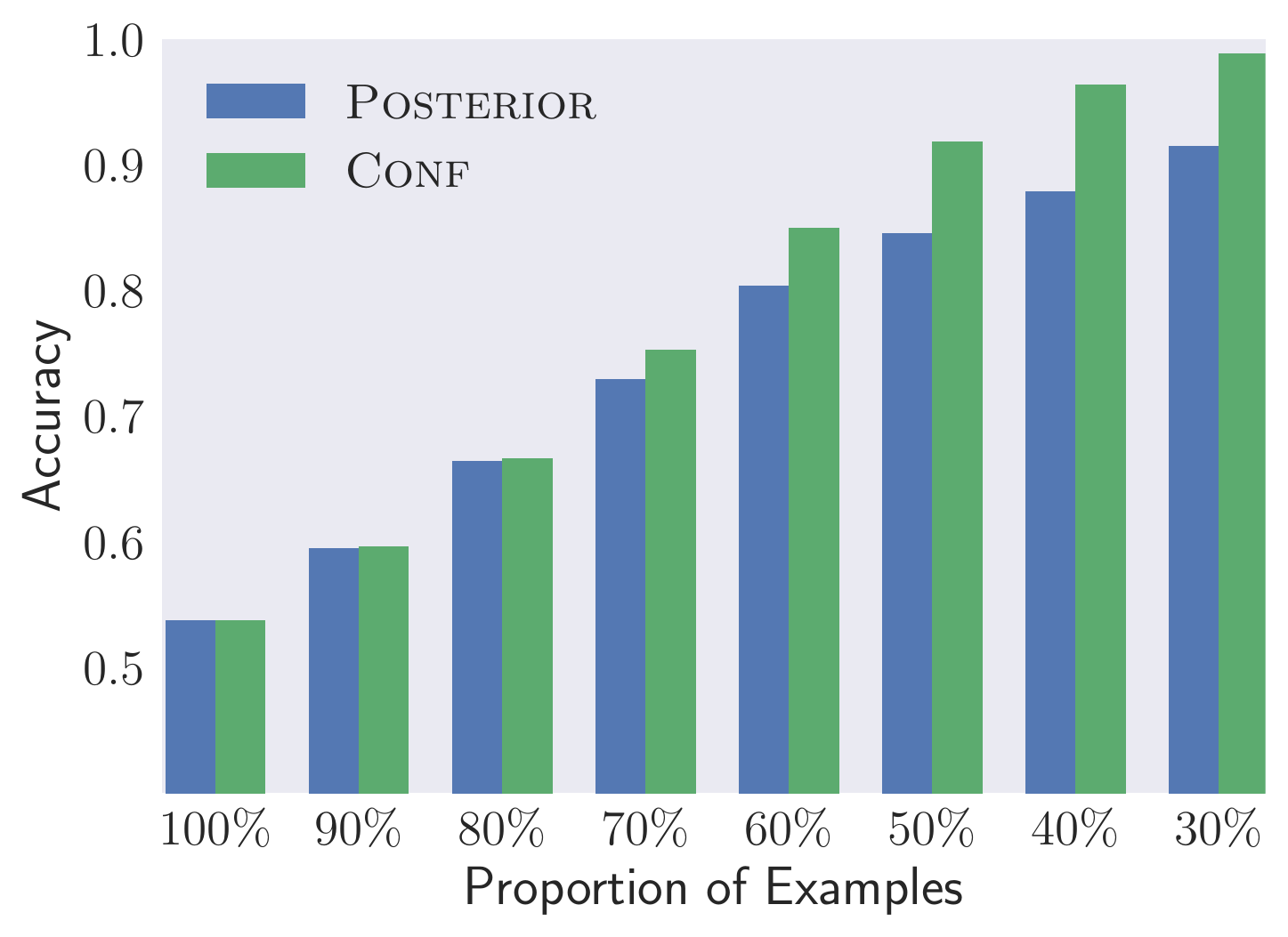}
	}
	%\vspace{-0.26cm}
	\caption{Confidence scores are used as threshold to filter out
		uncertain test examples. As the threshold increases,
		performance improves. The horizontal axis shows the proportion of
		examples beyond the threshold.}
	\label{fig:threshold}
\end{figure}

Finally, in real-world applications, confidence scores are often used
as a threshold to trade-off precision for coverage.
Figure~\ref{fig:threshold} shows how F1 score varies as we increase the
confidence threshold, i.e., reduce the proportion of examples that we
return answers for. F1 score improves monotonically for \textsc{Posterior}
and our method, which, however, achieves better performance when
coverage is the same.

\paragraph{Uncertainty Interpretation}
\label{sec:exp:interpretation}

We next evaluate how our backpropagation method (see
Section~\ref{sec:method:interpret}) allows us to identify input tokens
contributing to uncertainty.  We compare against a method that
interprets uncertainty based on the attention mechanism
(\textsc{Attention}). As shown in Equation~\eqref{eq:attention:score},
attention scores ${ r }_{ t,k }$ can be used as soft alignments
between the time step~$t$ of the decoder and the \mbox{$k$-th}~input
token. We compute the normalized uncertainty score~$\hat{u}_{q_t}$ for
a token~$q_t$ via:
\begin{equation}
\hat{u}_{q_t} \propto \sum_{t=1}^{|a|} { { r }_{ t,k } u_{a_t} }
\end{equation}
where~$u_{a_t}$ is the uncertainty score of the predicted token~$a_t$
(Equation~\eqref{eq:token-uncertainty}), and $\sum _{ t=1 }^{ |q| }{ \hat{u}_{q_t} } = 1$.

Unfortunately, the evaluation of uncertainty interpretation methods is
problematic. For our semantic parsing task, we do not a priori know
which tokens in the natural language input contribute to uncertainty
and these may vary depending on the architecture used, model
parameters, and so on. We work around this problem by creating a proxy
gold standard. We inject noise to the vectors representing tokens in
the encoder (see Section~\ref{sec:method:metric-model}) and then
estimate the uncertainty caused by each token~$q_t$
(Equation~\eqref{eq:dropout-uncertainty-seq}) under the assumption
that addition of noise should only affect genuinely uncertain
tokens. Notice that here we inject noise to one token at a
time\footnote{Noise injection as described above is used for
evaluation purposes only since we need to perform forward passes
multiple times (see Section~\ref{sec:method:metric-model}) for each
token, and 
the running time increases linearly with the input length.
}
instead of all parameters (see Figure~\ref{fig:dropout_where}).
Tokens identified as uncertain by the above procedure
are considered gold standard and
compared to those identified by our method. We use Gaussian noise to
perturb vectors in our experiments (dropout obtained similar
results).

We define an evaluation metric based on the overlap ($overlap@K$)
among tokens identified as uncertain by the model and the gold
standard.  Given an example, we first compute the interpretation
scores of the input tokens according to our method, and obtain a
list~$\tau_1$ of $K$~tokens with highest scores. We also obtain a list
$\tau_2$ of~$K$ tokens with highest ground-truth scores and measure
the degree of overlap between these two lists:
\begin{equation}
\text{overlap@}K = \frac{| \tau_1 \cap \tau_2 |}{ K } \nonumber
\end{equation}
where $K \in \{2,4\}$ in our experiments.
For example, the $\text{overlap@}4$ metric of the lists
$\tau_1 = [ q_7 , q_8 , q_2 , q_3 ]$ and
$\tau_2 = [ q_7 , q_8 , q_3 , q_4 ]$ is $3 / 4$, because there are three
overlapping tokens.

\begin{table}[t]
	\centering
	\small
	\begin{tabular}{l c c c c}
		\toprule
		\textbf{Method}    & \multicolumn{2}{c}{\textsc{\textbf{Ifttt}}} & \multicolumn{2}{c}{\textsc{\textbf{Django}}} \\
		\cmidrule{2-5}     & \textbf{@2}    & \textbf{@4}                & \textbf{@2}    & \textbf{@4}                 \\ \midrule
		\textsc{Attention} & 0.525          & 0.737                      & 0.637          & 0.684                       \\
		\textsc{BackProp}  & \textbf{0.608} & \textbf{0.791}             & \textbf{0.770} & \textbf{0.788}              \\ \bottomrule
	\end{tabular}
	\normalsize
	\caption{Uncertainty interpretation against
		inferred ground truth; we compute the overlap between tokens
		identified as contributing to uncertainty by our method and
		those found in the gold standard. Overlap is shown for top~$2$
		and~$4$ tokens. Best results are in \textbf{bold}.} 
	\label{table:results:interpret}
\end{table}

\begin{table}[t]
\centering
\small
\renewcommand{\tabcolsep}{2pt}
\begin{tabular}{@{}l p{7.3cm}@{}}
\toprule
% 1195
\multicolumn{2}{@{}l@{}}{\tabincell{l}{~ \\ ~ \\ ~}}
\varwidth{\linewidth}
\begin{lstlisting}[style=ifttt]
google_calendar-|\rg{any\_event\_starts}| THEN facebook-create_a_status_message-(status_message|\rg{(\{description\}}|))
\end{lstlisting}
\endvarwidth\hfill \\
\textsc{Att} & post calendar event \bg{to} facebook
\\
\textsc{BP} & post \bg{calendar event} to facebook
\\ \hline
\multicolumn{2}{l}{\tabincell{l}{~ \\ ~}}
\varwidth{\linewidth}
\begin{lstlisting}[style=ifttt]
feed-new_feed_item-(feed_url(|\rg{\_url\_sports.espn.go.com}|)) THEN |\textrm{...}|
\end{lstlisting}
\endvarwidth\hfill \\
\textsc{Att} & espn mlb \bg{headline to} readability
\\
\textsc{BP} & \bg{espn} mlb \bg{headline} to readability
\\ \hline
\multicolumn{2}{l}{\tabincell{l}{~ \\ ~}}
\varwidth{\linewidth}
\begin{lstlisting}[style=ifttt]
weather-|\rg{tomorrow's\_low\_drops\_below}|-((temperature(|\rg{0}|)) (degrees_in(c))) THEN |\textrm{...}|
\end{lstlisting}
\endvarwidth\hfill \\
\textsc{Att} & warn me when \bg{it's} going to be freezing \bg{tomorrow}
\\
\textsc{BP} & warn me when it's going to be \bg{freezing tomorrow}
\\ \hline
\multicolumn{2}{l}{\tabincell{l}{~}}
\varwidth{\linewidth}
\begin{lstlisting}[style=django]
if |\rg{str\_number}|[|\rg{0}|] == '_STR_':
\end{lstlisting}
\endvarwidth\hfill \\
\textsc{Att} & if first element \bg{of str\_number} equals a string \_STR\_ .
\\
\textsc{BP} & if \bg{first element of str\_number} equals a string \_STR\_ .
\\ \hline
\multicolumn{2}{l}{\tabincell{l}{~}}
\varwidth{\linewidth}
\begin{lstlisting}[style=django]
|\rg{start}| = 0
\end{lstlisting}
\endvarwidth\hfill \\
\textsc{Att} & start \bg{is} an integer 0 .
\\
\textsc{BP} & \bg{start} is an integer 0 .
\\ \hline
\multicolumn{2}{l}{\tabincell{l}{~}}
\varwidth{\linewidth}
\begin{lstlisting}[style=django]
if name.|\rg{startswith}|(|\rg{'}|_STR_'):
\end{lstlisting}
\endvarwidth\hfill \\
\textsc{Att} & if name starts \bg{with} an string \bg{\_STR\_} ,
\\
\textsc{BP} & if name \bg{starts with} an \bg{string \_STR\_} ,
\\
\bottomrule
\end{tabular}
\normalsize
%\vspace{-1ex}
\caption{Uncertainty interpretation for \textsc{Attention}
(\textsc{Att}) and \textsc{Backprop} (\textsc{BP})
%taken from \textsc{Ifttt} and \textsc{Django}
. The first line in each group is the model prediction. 
Predicted tokens and input words with large scores
are shown in \rg{red} and
\bg{blue}, respectively.}
\label{table:interpretation:examples}
\end{table}

Table~\ref{table:results:interpret} reports results with overlap@$2$
and overlap@$4$. Overall, \textsc{BackProp} achieves better
interpretation quality than the attention mechanism. On both datasets,
about $80\%$ of the \mbox{top-$4$} tokens identified as uncertain
agree with the ground truth. Table~\ref{table:interpretation:examples}
shows examples where our method has identified input tokens
contributing to the uncertainty of the output. We highlight
token~$a_t$ if its uncertainty score $u_{a_t}$ is greater than
\mbox{$0.5 * \avg \{ u_{a_{t'}} \}_{t'=1}^{|a|}$}.
The results illustrate that the parser tends to be uncertain about
tokens which are function arguments (e.g.,~URLs, and message content),
and ambiguous inputs. The examples show that \textsc{Backprop} is
qualitatively better compared to \textsc{Attention}; attention scores
often produce inaccurate alignments while \textsc{Backprop} can
utilize information flowing through the LSTMs rather than only relying
on the attention mechanism.

\section{Conclusions}
\label{sec:conclusions}

In this paper we presented a confidence estimation model and an
uncertainty interpretation method for neural semantic parsing.
Experimental results show that our method achieves better performance than competitive baselines on two datasets.
Directions for future work are many and varied.
The proposed framework could be applied to a
variety of tasks~\cite{mt:jointly:align:translate,grammar:correction} employing sequence-to-sequence architectures.
We could also utilize the confidence estimation model within an active learning framework for neural semantic parsing.

\section*{Acknowledgments}

We would like to thank Pengcheng Yin for sharing with us the
preprocessed version of the \textsc{Django} dataset.  We gratefully
acknowledge the financial support of the European Research Council
(award number 681760; Dong, Lapata) and the AdeptMind Scholar
Fellowship program (Dong).

\bibliographystyle{acl_natbib}
\bibliography{confidence}

\begin{thebibliography}{46}
\expandafter\ifx\csname natexlab\endcsname\relax\def\natexlab#1{#1}\fi

\bibitem[{Andreas et~al.(2013)Andreas, Vlachos, and Clark}]{sp:as:mt}
Jacob Andreas, Andreas Vlachos, and Stephen Clark. 2013.
\newblock \href {http://www.aclweb.org/anthology/P13-2009} {Semantic parsing as
  machine translation}.
\newblock In \emph{Proceedings of the 51st Annual Meeting of the Association
  for Computational Linguistics}, pages 47--52, Sofia, Bulgaria.

\bibitem[{Bach et~al.(2015)Bach, Binder, Montavon, Klauschen, Müller, and
  Samek}]{lrp}
Sebastian Bach, Alexander Binder, Grégoire Montavon, Frederick Klauschen,
  Klaus-Robert Müller, and Wojciech Samek. 2015.
\newblock \href {https://doi.org/10.1371/journal.pone.0130140} {On pixel-wise
  explanations for non-linear classifier decisions by layer-wise relevance
  propagation}.
\newblock \emph{PLOS ONE}, 10(7):1--46.

\bibitem[{Bahdanau et~al.(2015)Bahdanau, Cho, and
  Bengio}]{mt:jointly:align:translate}
Dzmitry Bahdanau, Kyunghyun Cho, and Yoshua Bengio. 2015.
\newblock \href {https://arxiv.org/pdf/1409.0473.pdf} {Neural machine
  translation by jointly learning to align and translate}.
\newblock In \emph{Proceedings of the 3rd International Conference on Learning
  Representations}, San Diego, California.

\bibitem[{Bird et~al.(2009)Bird, Klein, and Loper}]{nltk}
Steven Bird, Ewan Klein, and Edward Loper. 2009.
\newblock \href {http://www.nltk.org/book} {\emph{Natural Language Processing
  with Python}}.
\newblock O'Reilly Media.

\bibitem[{Blatz et~al.(2004)Blatz, Fitzgerald, Foster, Gandrabur, Goutte,
  Kulesza, Sanchis, and Ueffing}]{mt-confidence:coling}
John Blatz, Erin Fitzgerald, George Foster, Simona Gandrabur, Cyril Goutte,
  Alex Kulesza, Alberto Sanchis, and Nicola Ueffing. 2004.
\newblock \href {https://doi.org/10.3115/1220355.1220401} {Confidence
  estimation for machine translation}.
\newblock In \emph{Proceedings of the 20th International Conference on
  Computational Linguistics}, pages 315--321, Geneva, Switzerland.

\bibitem[{Blundell et~al.(2015)Blundell, Cornebise, Kavukcuoglu, and
  Wierstra}]{bbq}
Charles Blundell, Julien Cornebise, Koray Kavukcuoglu, and Daan Wierstra. 2015.
\newblock \href {http://dl.acm.org/citation.cfm?id=3045118.3045290} {Weight
  uncertainty in neural networks}.
\newblock In \emph{Proceedings of the 32nd International Conference on
  International Conference on Machine Learning}, pages 1613--1622, Lille,
  France.

\bibitem[{Chen and Guestrin(2016)}]{xgboost}
Tianqi Chen and Carlos Guestrin. 2016.
\newblock \href {https://doi.org/10.1145/2939672.2939785} {{XGBoost}: A
  scalable tree boosting system}.
\newblock In \emph{Proceedings of the 22nd ACM SIGKDD International Conference
  on Knowledge Discovery and Data Mining}, pages 785--794, San Francisco,
  California.

\bibitem[{Denker and Lecun(1991)}]{nn-bayesian91}
John~S Denker and Yann Lecun. 1991.
\newblock \href {http://dl.acm.org/citation.cfm?id=118850.119959} {Transforming
  neural-net output levels to probability distributions}.
\newblock In \emph{Advances in neural information processing systems}, pages
  853--859, Denver, Colorado.

\bibitem[{Dong and Lapata(2016)}]{lang2logic}
Li~Dong and Mirella Lapata. 2016.
\newblock \href {http://www.aclweb.org/anthology/P16-1004} {Language to logical
  form with neural attention}.
\newblock In \emph{Proceedings of the 54th Annual Meeting of the Association
  for Computational Linguistics}, pages 33--43, Berlin, Germany.

\bibitem[{Dong and Lapata(2018)}]{coarse2fine}
Li~Dong and Mirella Lapata. 2018.
\newblock Coarse-to-fine decoding for neural semantic parsing.
\newblock In \emph{Proceedings of the 56th Annual Meeting of the Association
  for Computational Linguistics}, Melbourne, Australia.

\bibitem[{Efron and Tibshirani(1994)}]{bootstrap}
Bradley Efron and Robert~J Tibshirani. 1994.
\newblock \emph{An Introduction to the Bootstrap}.
\newblock CRC press.

\bibitem[{Fan et~al.(2017)Fan, Monti, Mathias, and Dreyer}]{transfer-nsp}
Xing Fan, Emilio Monti, Lambert Mathias, and Markus Dreyer. 2017.
\newblock \href {http://www.aclweb.org/anthology/W17-2607} {Transfer learning
  for neural semantic parsing}.
\newblock In \emph{Proceedings of the 2nd Workshop on Representation Learning
  for {NLP}}, pages 48--56, Vancouver, Canada.

\bibitem[{Gal and Ghahramani(2016)}]{dropout-bayesian}
Yarin Gal and Zoubin Ghahramani. 2016.
\newblock \href {http://dl.acm.org/citation.cfm?id=3045390.3045502} {Dropout as
  a bayesian approximation: Representing model uncertainty in deep learning}.
\newblock In \emph{Proceedings of the 33rd International Conference on Machine
  Learning}, pages 1050--1059, New York City, NY.

\bibitem[{Gan et~al.(2017)Gan, Li, Chen, Pu, Su, and Carin}]{rnn-bayesian}
Zhe Gan, Chunyuan Li, Changyou Chen, Yunchen Pu, Qinliang Su, and Lawrence
  Carin. 2017.
\newblock \href {https://doi.org/10.18653/v1/P17-1030} {Scalable bayesian
  learning of recurrent neural networks for language modeling}.
\newblock In \emph{Proceedings of the 55th Annual Meeting of the Association
  for Computational Linguistics}, pages 321--331, Vancouver, Canada.

\bibitem[{Gondek et~al.(2012)Gondek, Lally, Kalyanpur, Murdock, Duboue, Zhang,
  Pan, Qiu, and Welty}]{watson}
D.~C. Gondek, A.~Lally, A.~Kalyanpur, J.~W. Murdock, P.~A. Duboue, L.~Zhang,
  Y.~Pan, Z.~M. Qiu, and C.~Welty. 2012.
\newblock \href {https://doi.org/10.1147/JRD.2012.2188760} {A framework for
  merging and ranking of answers in {DeepQA}}.
\newblock \emph{IBM Journal of Research and Development}, 56(3.4):14:1--14:12.

\bibitem[{Heafield et~al.(2013)Heafield, Pouzyrevsky, Clark, and Koehn}]{kenlm}
Kenneth Heafield, Ivan Pouzyrevsky, Jonathan~H. Clark, and Philipp Koehn. 2013.
\newblock \href {https://kheafield.com/papers/edinburgh/estimate\_paper.pdf}
  {Scalable modified {Kneser-Ney} language model estimation}.
\newblock In \emph{Proceedings of the 51st Annual Meeting of the Association
  for Computational Linguistics}, pages 690--696, Sofia, Bulgaria.

\bibitem[{Herzig and Berant(2017)}]{multi-kb-nsp}
Jonathan Herzig and Jonathan Berant. 2017.
\newblock \href {https://doi.org/10.18653/v1/P17-2098} {Neural semantic parsing
  over multiple knowledge-bases}.
\newblock In \emph{Proceedings of the 55th Annual Meeting of the Association
  for Computational Linguistics}, pages 623--628, Vancouver, Canada.

\bibitem[{Hochreiter and Schmidhuber(1997)}]{lstm}
Sepp Hochreiter and J\"{u}rgen Schmidhuber. 1997.
\newblock \href {https://doi.org/10.1162/neco.1997.9.8.1735} {Long short-term
  memory}.
\newblock \emph{Neural Computation}, 9:1735--1780.

\bibitem[{Iyer et~al.(2017)Iyer, Konstas, Cheung, Krishnamurthy, and
  Zettlemoyer}]{user-feedback}
Srinivasan Iyer, Ioannis Konstas, Alvin Cheung, Jayant Krishnamurthy, and Luke
  Zettlemoyer. 2017.
\newblock \href {http://aclweb.org/anthology/P17-1089} {Learning a neural
  semantic parser from user feedback}.
\newblock In \emph{Proceedings of the 55th Annual Meeting of the Association
  for Computational Linguistics}, pages 963--973, Vancouver, Canada.

\bibitem[{Jia and Liang(2016)}]{data-recombination}
Robin Jia and Percy Liang. 2016.
\newblock \href {http://www.aclweb.org/anthology/P16-1002} {Data recombination
  for neural semantic parsing}.
\newblock In \emph{Proceedings of the 54th Annual Meeting of the Association
  for Computational Linguistics}, pages 12--22, Berlin, Germany.

\bibitem[{Johansen and Socher(2017)}]{learn-skim-read}
Alexander Johansen and Richard Socher. 2017.
\newblock \href {http://www.aclweb.org/anthology/W17-2631} {Learning when to
  skim and when to read}.
\newblock In \emph{Proceedings of the 2nd Workshop on Representation Learning
  for {NLP}}, pages 257--264, Vancouver, Canada.

\bibitem[{Ko\v{c}isk\'{y} et~al.(2016)Ko\v{c}isk\'{y}, Melis, Grefenstette,
  Dyer, Ling, Blunsom, and Hermann}]{semi-nsp}
Tom\'{a}\v{s} Ko\v{c}isk\'{y}, G\'{a}bor Melis, Edward Grefenstette, Chris
  Dyer, Wang Ling, Phil Blunsom, and Karl~Moritz Hermann. 2016.
\newblock \href {https://aclweb.org/anthology/D16-1116} {Semantic parsing with
  semi-supervised sequential autoencoders}.
\newblock In \emph{Proceedings of the 2016 Conference on Empirical Methods in
  Natural Language Processing}, pages 1078--1087, Austin, Texas.

\bibitem[{Krishnamurthy et~al.(2017)Krishnamurthy, Dasigi, and
  Gardner}]{table-nsp}
Jayant Krishnamurthy, Pradeep Dasigi, and Matt Gardner. 2017.
\newblock \href {https://www.aclweb.org/anthology/D17-1160} {Neural semantic
  parsing with type constraints for semi-structured tables}.
\newblock In \emph{Proceedings of the 2017 Conference on Empirical Methods in
  Natural Language Processing}, pages 1517--1527, Copenhagen, Denmark.

\bibitem[{Kwiatkowski et~al.(2011)Kwiatkowski, Zettlemoyer, Goldwater, and
  Steedman}]{fubl}
Tom Kwiatkowski, Luke Zettlemoyer, Sharon Goldwater, and Mark Steedman. 2011.
\newblock \href {http://www.aclweb.org/anthology/D11-1140} {Lexical
  generalization in {CCG} grammar induction for semantic parsing}.
\newblock In \emph{Proceedings of the 2011 Conference on Empirical Methods in
  Natural Language Processing}, pages 1512--1523, Edinburgh, Scotland.

\bibitem[{Ling et~al.(2016)Ling, Blunsom, Grefenstette, Hermann,
  Ko\v{c}isk\'{y}, Wang, and Senior}]{latent-predictor}
Wang Ling, Phil Blunsom, Edward Grefenstette, Karl~Moritz Hermann,
  Tom\'{a}\v{s} Ko\v{c}isk\'{y}, Fumin Wang, and Andrew Senior. 2016.
\newblock \href {http://www.aclweb.org/anthology/P16-1057} {Latent predictor
  networks for code generation}.
\newblock In \emph{Proceedings of the 54th Annual Meeting of the Association
  for Computational Linguistics}, pages 599--609, Berlin, Germany.

\bibitem[{Lu et~al.(2008)Lu, Ng, Lee, and Zettlemoyer}]{lnlz08}
Wei Lu, Hwee~Tou Ng, Wee~Sun Lee, and Luke Zettlemoyer. 2008.
\newblock \href {http://www.aclweb.org/anthology/D08-1082} {A generative model
  for parsing natural language to meaning representations}.
\newblock In \emph{Proceedings of the 2008 Conference on Empirical Methods in
  Natural Language Processing}, pages 783--792, Honolulu, Hawaii.

\bibitem[{Luong et~al.(2015{\natexlab{a}})Luong, Pham, and
  Manning}]{luong-attention}
Thang Luong, Hieu Pham, and Christopher~D. Manning. 2015{\natexlab{a}}.
\newblock \href {http://aclweb.org/anthology/D15-1166} {Effective approaches to
  attention-based neural machine translation}.
\newblock In \emph{Proceedings of the 2015 Conference on Empirical Methods in
  Natural Language Processing}, pages 1412--1421, Lisbon, Portugal.

\bibitem[{Luong et~al.(2015{\natexlab{b}})Luong, Sutskever, Le, Vinyals, and
  Zaremba}]{mt:rare:word:google}
Thang Luong, Ilya Sutskever, Quoc Le, Oriol Vinyals, and Wojciech Zaremba.
  2015{\natexlab{b}}.
\newblock \href {http://www.aclweb.org/anthology/P15-1002} {Addressing the rare
  word problem in neural machine translation}.
\newblock In \emph{Proceedings of the 53rd Annual Meeting of the Association
  for Computational Linguistics and the 7th International Joint Conference on
  Natural Language Processing}, pages 11--19, Beijing, China.

\bibitem[{MacKay(1992)}]{nn-bayesian92}
David J.~C. MacKay. 1992.
\newblock \href {https://doi.org/10.1162/neco.1992.4.3.448} {A practical
  bayesian framework for backpropagation networks}.
\newblock \emph{Neural Computation}, 4(3):448--472.

\bibitem[{Neal(1996)}]{nn-bayesian96}
Radford~M Neal. 1996.
\newblock \href {https://doi.org/10.1007/978-1-4612-0745-0} {\emph{Bayesian
  learning for neural networks}}, volume 118.
\newblock Springer Science \& Business Media.

\bibitem[{Oda et~al.(2015)Oda, Fudaba, Neubig, Hata, Sakti, Toda, and
  Nakamura}]{django}
Yusuke Oda, Hiroyuki Fudaba, Graham Neubig, Hideaki Hata, Sakriani Sakti,
  Tomoki Toda, and Satoshi Nakamura. 2015.
\newblock \href {https://doi.org/10.1109/ASE.2015.36} {Learning to generate
  pseudo-code from source code using statistical machine translation}.
\newblock In \emph{Proceedings of the 2015 30th IEEE/ACM International
  Conference on Automated Software Engineering}, pages 574--584, Washington,
  DC.

\bibitem[{Quirk et~al.(2015)Quirk, Mooney, and Galley}]{ifttt}
Chris Quirk, Raymond Mooney, and Michel Galley. 2015.
\newblock \href {http://www.aclweb.org/anthology/P15-1085} {Language to code:
  Learning semantic parsers for if-this-then-that recipes}.
\newblock In \emph{Proceedings of the 53rd Annual Meeting of the Association
  for Computational Linguistics and the 7th International Joint Conference on
  Natural Language Processing}, pages 878--888, Beijing, China.

\bibitem[{Rabinovich et~al.(2017)Rabinovich, Stern, and Klein}]{asn}
Maxim Rabinovich, Mitchell Stern, and Dan Klein. 2017.
\newblock \href {http://aclweb.org/anthology/P17-1105} {Abstract syntax
  networks for code generation and semantic parsing}.
\newblock In \emph{Proceedings of the 55th Annual Meeting of the Association
  for Computational Linguistics}, pages 1139--1149, Vancouver, Canada.

\bibitem[{Schmaltz et~al.(2017)Schmaltz, Kim, Rush, and
  Shieber}]{grammar:correction}
Allen Schmaltz, Yoon Kim, Alexander Rush, and Stuart Shieber. 2017.
\newblock \href {https://www.aclweb.org/anthology/D17-1297} {Adapting sequence
  models for sentence correction}.
\newblock In \emph{Proceedings of the 2017 Conference on Empirical Methods in
  Natural Language Processing}, pages 2797--2803, Copenhagen, Denmark.

\bibitem[{Soricut and Echihabi(2010)}]{mt-confidence:acl}
Radu Soricut and Abdessamad Echihabi. 2010.
\newblock \href {http://www.aclweb.org/anthology/P10-1063} {Trustrank: Inducing
  trust in automatic translations via ranking}.
\newblock In \emph{Proceedings of the 48th Annual Meeting of the Association
  for Computational Linguistics}, pages 612--621, Uppsala, Sweden.

\bibitem[{Srivastava et~al.(2014)Srivastava, Hinton, Krizhevsky, Sutskever, and
  Salakhutdinov}]{dropout}
Nitish Srivastava, Geoffrey Hinton, Alex Krizhevsky, Ilya Sutskever, and Ruslan
  Salakhutdinov. 2014.
\newblock \href {http://jmlr.org/papers/v15/srivastava14a.html} {Dropout: A
  simple way to prevent neural networks from overfitting}.
\newblock \emph{Journal of Machine Learning Research}, 15:1929--1958.

\bibitem[{Susanto and Lu(2017)}]{multilingual-nsp}
Raymond~Hendy Susanto and Wei Lu. 2017.
\newblock \href {http://aclweb.org/anthology/P17-2007} {Neural architectures
  for multilingual semantic parsing}.
\newblock In \emph{Proceedings of the 55th Annual Meeting of the Association
  for Computational Linguistics}, pages 38--44, Vancouver, Canada.

\bibitem[{Sutskever et~al.(2014)Sutskever, Vinyals, and Le}]{mt:seq2seq}
Ilya Sutskever, Oriol Vinyals, and Quoc~V Le. 2014.
\newblock \href {http://dl.acm.org/citation.cfm?id=2969033.2969173} {Sequence
  to sequence learning with neural networks}.
\newblock In \emph{Advances in Neural Information Processing Systems}, pages
  3104--3112, Montreal, Canada.

\bibitem[{Tang and Mooney(2000)}]{tang-mooney:2000:EMNLP}
Lappoon~R. Tang and Raymond~J. Mooney. 2000.
\newblock \href {https://doi.org/10.3115/1117794.1117811} {Automated
  construction of database interfaces: Intergrating statistical and relational
  learning for semantic parsing}.
\newblock In \emph{2000 Joint SIGDAT Conference on Empirical Methods in Natural
  Language Processing and Very Large Corpora}, pages 133--141, Hong Kong,
  China.

\bibitem[{Tieleman and Hinton(2012)}]{rmsprop}
T.~Tieleman and G.~Hinton. 2012.
\newblock {Lecture 6.5---{RMSProp}: Divide the gradient by a running average of
  its recent magnitude}.
\newblock Technical report.

\bibitem[{Ueffing and Ney(2005)}]{mt-confidence:cl}
Nicola Ueffing and Hermann Ney. 2005.
\newblock \href {https://doi.org/10.3115/1220575.1220671} {Word-level
  confidence estimation for machine translation using phrase-based translation
  models}.
\newblock In \emph{Proceedings of the Conference on Human Language Technology
  and Empirical Methods in Natural Language Processing}, pages 763--770,
  Vancouver, Canada.

\bibitem[{Xiao et~al.(2016)Xiao, Dymetman, and Gardent}]{grammar-nsp}
Chunyang Xiao, Marc Dymetman, and Claire Gardent. 2016.
\newblock \href {http://www.aclweb.org/anthology/P16-1127} {Sequence-based
  structured prediction for semantic parsing}.
\newblock In \emph{Proceedings of the 54th Annual Meeting of the Association
  for Computational Linguistics}, pages 1341--1350, Berlin, Germany.

\bibitem[{Yin and Neubig(2017)}]{nl2code}
Pengcheng Yin and Graham Neubig. 2017.
\newblock \href {http://aclweb.org/anthology/P17-1041} {A syntactic neural
  model for general-purpose code generation}.
\newblock In \emph{Proceedings of the 55th Annual Meeting of the Association
  for Computational Linguistics}, pages 440--450, Vancouver, Canada.

\bibitem[{Zettlemoyer and Collins(2007)}]{zc07}
Luke Zettlemoyer and Michael Collins. 2007.
\newblock \href {http://www.aclweb.org/anthology/D07-1071} {Online learning of
  relaxed {CCG} grammars for parsing to logical form}.
\newblock In \emph{Proceedings of the 2007 Joint Conference on Empirical
  Methods in Natural Language Processing and Computational Natural Language
  Learning}, pages 678--687, Prague, Czech Republic.

\bibitem[{Zhang et~al.(2016)Zhang, Lin, Brandt, Shen, and
  Sclaroff}]{excitation}
Jianming Zhang, Zhe Lin, Jonathan Brandt, Xiaohui Shen, and Stan Sclaroff.
  2016.
\newblock \href {https://doi.org/10.1007/s11263-017-1059-x} {Top-down neural
  attention by excitation backprop}.
\newblock In \emph{European Conference on Computer Vision}, pages 543--559,
  Amsterdam, Netherlands.

\bibitem[{Zhao and Huang(2015)}]{tisp}
Kai Zhao and Liang Huang. 2015.
\newblock \href {http://www.aclweb.org/anthology/N15-1162} {Type-driven
  incremental semantic parsing with polymorphism}.
\newblock In \emph{Proceedings of the 2015 Conference of the North American
  Chapter of the Association for Computational Linguistics: Human Language
  Technologies}, pages 1416--1421, Denver, Colorado.

\end{thebibliography}
	
\end{document}